\documentclass[
twocolumn,
]{ceurart}

\usepackage{listings}

\usepackage[small]{caption}
\usepackage{csquotes}
\usepackage{subcaption}

\begin{document}

\copyrightyear{2022}
\copyrightclause{Copyright for this paper by its authors.
  Use permitted under Creative Commons License Attribution 4.0
  International (CC BY 4.0).}

\conference{The IJCAI-ECAI-22 Workshop on Artificial Intelligence Safety (AISafety 2022),
  July 24-25, 2022, Vienna, Austria}

\title{Leveraging generative models to characterize the failure conditions of image classifiers}

\author[1,2]{Adrien Le\ Coz}[%
email=adrien.le-coz@irt-systemx.fr,
]
\cormark[1]

\address[1]{IRT SystemX, Palaiseau, France}
\address[2]{DTIS, ONERA, Université Paris Saclay F-91123 Palaiseau - France}

\author[2]{Stéphane Herbin}[%
orcid=0000-0002-3341-3018,
email=stephane.herbin@onera.fr,
]
\cormark[1]

\author[1]{Faouzi Adjed}[
orcid=0000-0002-0100-9352,
email=faouzi.adjed@irt-systemx.fr
]
\cormark[1]

\cortext[1]{Corresponding author.}

\begin{abstract}
  We address in this work the question of identifying the failure conditions of a given image classifier. To do so, we exploit the capacity of producing controllable distributions of high quality image data made available by recent Generative Adversarial Networks (StyleGAN2): the failure conditions are expressed as directions of strong performance degradation in the generative model latent space. This strategy of analysis is used to discover corner cases that combine multiple sources of corruption, and to compare in more details the behavior of different classifiers. The directions of degradation can also be rendered visually by generating data for better interpretability. Some degradations such as image quality can affect all classes, whereas other ones such as shape are more class-specific. The approach is demonstrated on the MNIST dataset that has been completed by two sources of corruption: noise and blur, and shows a promising way to better understand and control the risks of exploiting Artificial Intelligence components for safety-critical applications.
\end{abstract}


\maketitle

\begin{figure*}[t]
\centering
\includegraphics[width=0.8\textwidth]{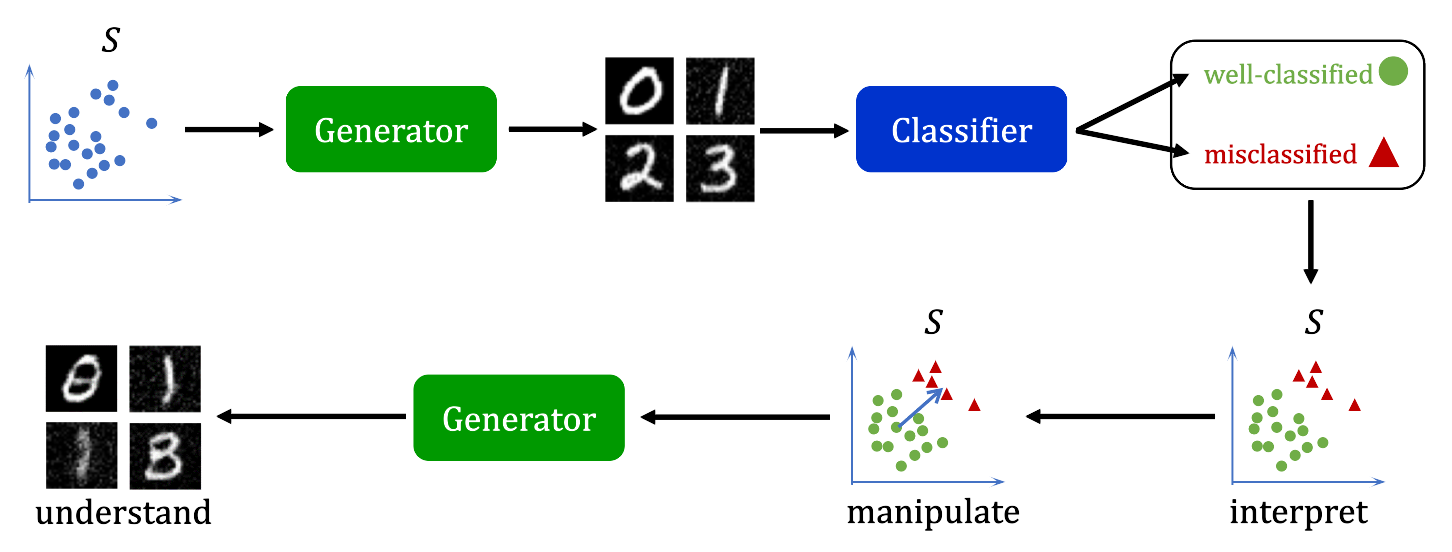}
\caption{\label{fig:diagram}An illustration of the approach. Starting from the latent space $\mathcal{S}$ of StyleGAN, we generate a population of images. The images are classified and the information on classification success is added in the space $\mathcal{S}$, where we find the dimensions discriminating well-classified vs. mis-classified images. These dimensions can then be used to visually render the corresponding influential attributes.
}
\end{figure*}

\section{Introduction}

\subsection*{Trustworthy AI}

Artificial Intelligence (AI) is getting every year more mature with potential applications to real world problems, and possibly to safety critical systems. Machine Learning (ML) is one of the most prominent set of AI techniques used to design predictive functions, especially for high dimensional inputs such as image, video, text or sound and generally involves Deep Neural Networks (DNN),

The exploitation of ML techniques introduces new issues to ensure safety and trustworthiness when designing or integrating AI based components: data quality assessment, robustness to adversarial perturbations, formal verification of DNNs, explainability, DNN calibration, etc. These research actions are complemented by the production of a large number of position papers and reports produced by academic, industrial and government organizations or working groups (ISO, SAE, NHTSA, EASA, HLEG of EU, DEEL, etc.); one of the main objectives of which is to renew certification standards so that the various phases of an industrial process (specification, design, validation \& verification, deployment, integration, operation, versioning, etc.) can accommodate AI.

In spite of all this on-going activity, the available design tools have difficulty to master with an acceptable level of trustworthiness the complexity of AI-ML components for real-world safety-critical applications. The work presented in this paper contributes to better understand the behavior of an AI-ML component, and  to identify the measurable and/or verifiable conditions influencing success or failure. Its long-term motivation is to close the loop between the specification, design and testing steps by providing more refined analytical tools. The target application domain is computer vision where AI-ML techniques are now ubiquitous.

\subsection*{Characterizing AI components}

Knowing on what conditions, or with what probability, a given component may fail is a key information for designing reliable systems.

When it comes to AI-ML, a classical approach to analyze a given algorithm is to build a test dataset, often a split from a large dataset, and compute performance indicators measuring the discrepancy between the ideal and actual predictions (accuracy, precision/recall, area under the ROC curve, etc.) 

A pure data-driven strategy to characterize the intended function and the good or undesired behaviors of an AI-ML component raises several issues. (1) most of the usual performance indicators are global statistics and cannot express in a fine grained way the algorithm behavior: they can be used to rank competing solutions -- this is currently done in academic benchmarks -- but are not able to identify what are their specific failure conditions, i.e. on what kind of input an algorithm is good or bad compared to others.
(2) it is difficult to gather all the good and bad operating conditions into one data set. There have been some attempts to describe rather exhaustively the possible hazards to families of algorithms \cite{zendel2017how}, but what these attempts in fact revealed was the complexity to master. Test dataset replication experiences have also shown that for high dimensional data, performance measures can have large variance \cite{engstrom2020identifying,recht2019imagenet}.
(3) typical causes of performance degradation of AI-ML components such as distributional shift \cite{koh2021wilds} and instability to small perturbations \cite{akhtar2021advances,li2020adversarial} are difficult to catch with a single test dataset.

Another approach to characterize a given component, inspired by software engineering practices, is to define a \emph{testing} strategy \textquote{designed to reveal machine learning bugs} \cite{zhang2022machine}. For instance, \cite{pei2017deepxplore} exploits a concept of neuron coverage inspired by test coverage in traditional software testing, to detect erroneous inputs.

In our approach, we propose to combine these two different strategies, data-driven evaluation and testing, in order to characterize the behavior of a given function: we identify the influential causes of performance degradation by evaluating the performance on sets of generated data that sample various data attributes, corruption or nuisance.

\subsection*{Generative models to explore data space}

Designing a probabilistic model in high dimensional data space such as image, video or sound, able to faithfully account for their diversity and informative features is a difficult (impossible?) objective. Generative models such as GANs \cite{GAN} or  generative invertible flows \cite{kingma2018glow} is a series of ML techniques that provides means to give access to such a distribution by direct sampling. What is learned is not the parameters of the probability density but the parameters of a sampling process able to generate data that mimic a given random distribution. 

GANs exploit a representational \emph{latent space} that can be sampled from a known low-dimension distribution, often Gaussian, that is expected to encode enough information to generate complete images. Generation is then produced by a decoding network that is learned from target data samples. Recent approaches \cite{wang2021generative,jabbar2021survey,saxena2021generative} are now able to generate high quality high dimension data, with a photo realistic rendering when applied to images, and with good diversity and fidelity levels. One possible application of generative models for safety objectives is to augment data for testing various operational conditions as in \cite{zhang2018deeproadb}.

The latent space can also be used as a way to control the generation process, for instance to edit images \cite{GANSpace,stylespace,EditGAN}. When correctly disentangled, the latent space can be interpreted as a representation space where each dimension encodes some interpretable visual attribute \cite{stylespace,StyleCLIP}. In the case of face image generation, these attributes could be hairstyle, head orientation, eye color, glasses, etc. Navigating in the representational latent space can also be used to identify the attributes that characterize best a given class \cite{explaining_in_style}. 


\subsection*{Main contributions}

We show how to exploit generative models to finely analyze the behavior of classifiers with high dimensional input in order to:
\begin{itemize}
\item identify influential directions of performance degradation that can be expressed both in the data space and in a latent feature space of a generative model;
\item discover \emph{corner cases} by exploring the directions of degradation in the latent feature space;
\item compare classifier performance on influential data features.
\end{itemize}
We focus in this paper on image classification as one of the paradigmatic decision problems of computer vision with object detection and semantic segmentation, and illustrate our method on a corrupted version of the MNIST dataset \cite{lecun2010mnist} .

\section{Proposed approach}
\label{sec:proposed_appraoch}

The proposed approach is illustrated in Figure~\ref{fig:diagram}, where we explore how the latent space of a generative model differentiates between data that are well and poorly classified by a given classifier. In the following, we will briefly describe the chosen generative model  and its latent space structure (Section \ref{sec:prerequisites}); explain how to find the dimensions of the latent space that differentiate well-classified from mis-classified data (Section \ref{sec:finding_dimensions}); describe how to manipulate images to visualize the attributes (Section \ref{sec:image_manipulation}); and see how we can estimate the accuracy of the classifier conditionally to the location of the data in the latent space (Section \ref{sec:accuracy}).

\begin{figure}[t]
\centering
\includegraphics[width=0.48\textwidth]{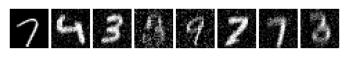}
\includegraphics[width=0.48\textwidth]{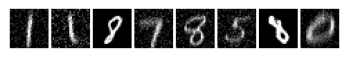}
 \caption{Samples of real corrupted data (top row) vs. generated data (bottom row)}
 \label{fig:samples_real_and_generated}
\end{figure}

\subsection{Resources}
\label{sec:prerequisites}
The current work is mainly based on two resources and objects that we present upstream in the two next subsections. They represent the theoretical and necessary tools allow us to better detail our proposed approach.   

\subsubsection{Classifier and data}
The first input to our approach is a learned image classifier to be analyzed. We assume that we have access to its architecture and weights (\textquote{white box}). We also overcome what is the domain of application (handwritten digits, faces, indoor scenes, etc.), and have a corresponding dataset available, not necessarily used for learning the classifier.

\subsubsection{Generative model}
The second ingredient of our approach is a generative model that can be controlled meaningfully.
In our work, we used the StyleGAN2 model \cite{stylegan2-ada} for a few reasons: the quality of generated data, its scalability to complex datasets, and the various levels of latent spaces. Indeed, three different latent spaces can be considered. The first latent space, $\mathcal{Z}$, is typically normally distributed like many GANs and is the initial input space of the generator. Samples $\mathbf{z} \in \mathcal{Z}$ are forwarded to the intermediate latent space $\mathcal{W}$ using fully connected layers, resulting in a more disentangled representation than $\mathcal{Z}$ \cite{stylegan_ffhq}.  Using learned affine transformations, samples $\mathbf{w} \in \mathcal{W}$ are specialized into \emph{styles} that scale the convolution weights for each feature map for each layer of the generator. A generated image is the result of an initial learned constant tensor that is up-sampled and transformed by residual convolution layers that are modulated by the style vector. Images are generated from the style vector $\mathbf{s}$ by the generator $G(\mathbf{s})$. The space of styles, called StyleSpace, shows a high degree of disentanglement \cite{stylespace}. This latent space $\mathcal{S}$ encodes distinct visual attributes along its dimensions and is typically used for image editing. 
To give an idea of the complexity of the generative model, in the original StyleGAN2 version that generates images of size $1024^2$, $\mathcal{Z}$ and $\mathcal{W}$ have $512$ dimensions, $\mathcal{S}$ has $9088$ dimensions, and the initial constant tensor has a size of $4^2$ with $512$ channels.

\begin{figure}[t]
    \centering
    \begin{subfigure}{0.33\linewidth}
        \includegraphics[width=\linewidth]{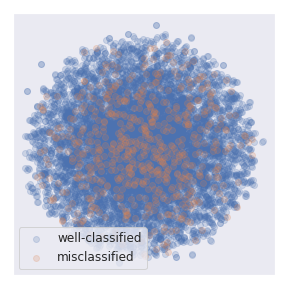}
        \caption{\label{fig:tsne_z}t-SNE in $\mathcal{Z}$}
    \end{subfigure}
    \begin{subfigure}{0.33\linewidth}
        \includegraphics[width=\linewidth]{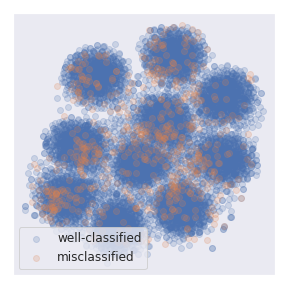}
        \caption{\label{fig:tsne_w}t-SNE in $\mathcal{W}$}
    \end{subfigure}%
    \begin{subfigure}{0.33\linewidth}
        \includegraphics[width=\linewidth]{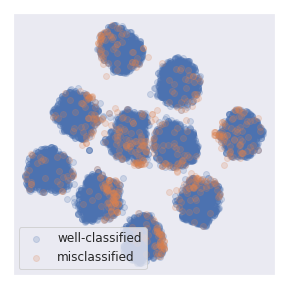}
        \caption{\label{fig:tsne_s}t-SNE in $\mathcal{S}$}
    \end{subfigure}%
    \caption{\label{fig:t-SNE}t-SNE of generated samples in different latent spaces. $\mathcal{Z}$ does not encode the class as class information is concatenated to latent codes $z$ to form the input of the conditional generator, and does not clearly differentiate well-classified from mis-classified samples. $\mathcal{W}$ and $\mathcal{S}$ are able to separate the classes (the 10 clusters) and well-classified from mis-classified samples, $\mathcal{S}$ doing it better than $\mathcal{W}$.}
\end{figure}

\subsection{Finding influential dimensions in the latent space}
\label{sec:finding_dimensions}
The dimensions of the latent StyleSpace $\mathcal{S}$ are expected to encode image attributes, such as shape, thickness, orientation and noise, in a rather disentangled way. We exploit this property to define a simple search method able to identify the most influential dimensions regarding the accuracy of a given classifier.

\paragraph{Gradient based approach}

The proposed strategy ranks the dimensions according to the gradient of the classifier output with respect to the StyleSpace input. The idea is to score each dimension based on its ability to lower the output score of the true class. More precisely, for each sample $\mathbf{s}$ in the StyleSpace, for which we know the true class, we generate the corresponding image $\mathbf{x}=G(\mathbf{s})$, and then classify it according to $C(\mathbf{x})$. Then we compute the gradient with respect to the dimension $j$ in the style space of the $i$-th classification output: $\nabla_{s_j} (C_i(G(\mathbf{s}))$, where $i$ is the index of the true class encoded by $\mathbf{s}$. The gradient can be computed exactly by using an \texttt{autograd} algorithmic differentiation provided in standard Deep Learning software environments -- both the classifier and the generator being available in such framework. We compute the average gradient over a population of data as the score used to rank the dimensions.

\paragraph{Global and class based analysis}

Not all classes behave similarly when corrupted. For instance, a $1$ digit, usually written as a single stroke, is more easily identified than a $3$, which can be mis-classified as an $8$ when there is noise. The impact of corruption potentially depends on the class. 

Our approach to compute influential directions rely on an average over a population. This population can be global or conditioned by the class, allowing a class conditional or global discovery of influential directions.




\subsection{Image manipulation and corner cases}
\label{sec:image_manipulation}
Starting from an image where the latent space representation -- the style vector -- is known, we can modify this representation to generate a modified image. In fact, once the influential dimensions are computed (see Section \ref{sec:finding_dimensions} above) and if 
we change the values of the style vector for those dimensions, then we modify the corresponding visual attributes for the generated image. Generating data that follow a high performance degradation is a simple heuristic: (1) we start from a given point $\mathbf{s_0}$ in the StyleSpace, (2) we increment the influential dimension by a given amount, and (3) we monitor the sign of the increment being given by the sign of the gradient. Note that the starting point $\mathbf{s_0}$ for exploration can be any point in the StyleSpace: it can be a \textquote{true} style, computed by mapping to $\mathcal{S}$ a random $\mathbf{z}$ sampled in the input latent space $\mathcal{Z}$, or any other point directly sampled in $\mathcal{S}$, for instance, an average of a given population of $\mathbf{s}$ data. We will use in the experiments (section~\ref{sec:results}) an \textquote{average} digit in the StyleSpace computed as the mean over a class conditional population.

This data space exploration along influential dimensions allows also the discovery of \emph{corner cases} defined as the smallest degradation that shifts the classifier output from good to bad classification. The experiments in~\ref{sec:results} will show several examples of corner cases discovered by this approach.


\subsection{Accuracy conditioned by latent space}
\label{sec:accuracy}
We can also use the latent space to understand better the classifier accuracy. In Section \ref{sec:finding_dimensions} we described how to find influential dimensions in the latent space. The data along those dimensions thus potentially correlates with the classifier accuracy. A classifier can be characterized globally by its accuracy decrease when fed by various amount of corruption produced by moving in the StyleSpace along influential dimensions. The decreasing slope characterizes globally the resilience to corruption of a classifier and can be used for comparison.

\section{Results}
\label{sec:results}

\begin{figure}[t]
    \centering
    \begin{subfigure}{0.48\textwidth}
        \centering
        \includegraphics[width=\textwidth]{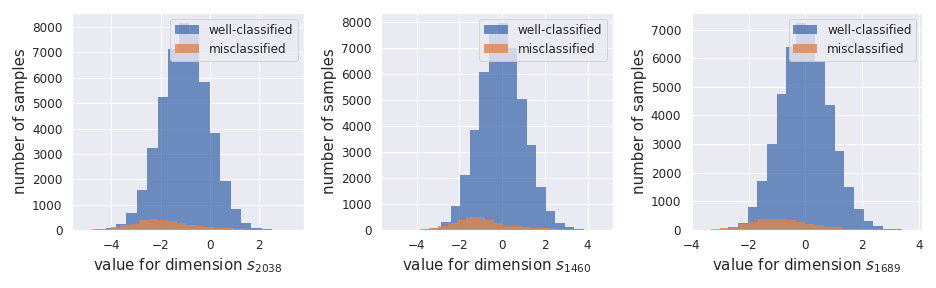}
        \caption{\label{fig:histograms_s_top}Top dimensions}
    \end{subfigure}
    \hfill
    \begin{subfigure}{0.48\textwidth}
        \centering
        \includegraphics[width=\textwidth]{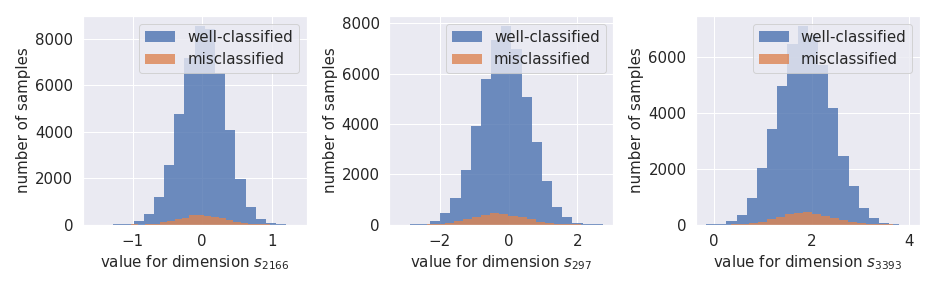}
        \caption{\label{fig:histograms_s_random}Random dimensions}
    \end{subfigure}
    \caption{\label{fig:histograms_s}(\subref{fig:histograms_s_top}) Histograms of values for the top 3 dimensions of $\mathcal{S}$ that discriminate the most between well-classified and mis-classified images after generation. For those top dimensions it is clear that latent codes resulting in well-classified and mis-classified images follow different distributions. \\
    (\subref{fig:histograms_s_random}) Histogram of values for 3 random dimensions of $\mathcal{S}$. For those dimensions, no difference is visible between the well-classified and mis-classified distributions.}
\end{figure}

\begin{figure}[t]
\centering
\includegraphics[width=0.48\textwidth]{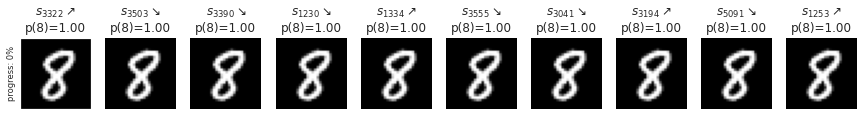}
\includegraphics[width=0.48\textwidth]{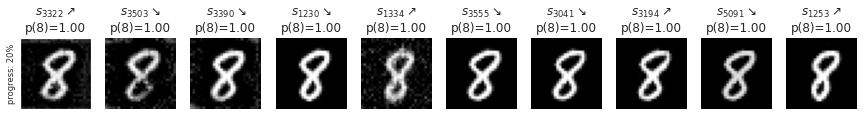}
\includegraphics[width=0.48\textwidth]{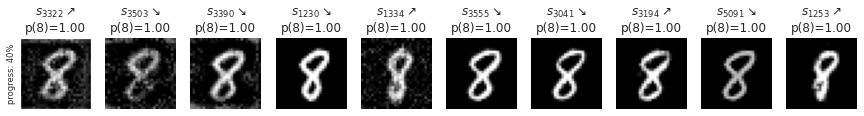}
\includegraphics[width=0.48\textwidth]{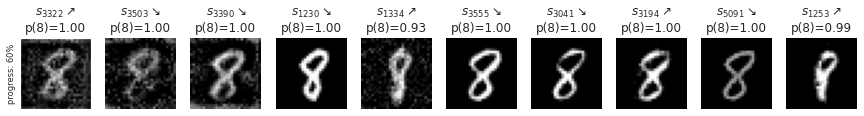}
\includegraphics[width=0.48\textwidth]{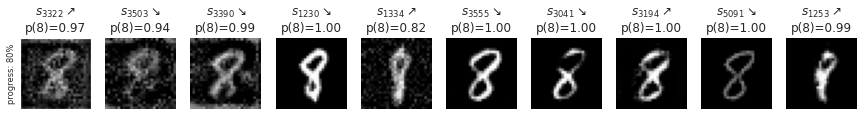}
\includegraphics[width=0.48\textwidth]{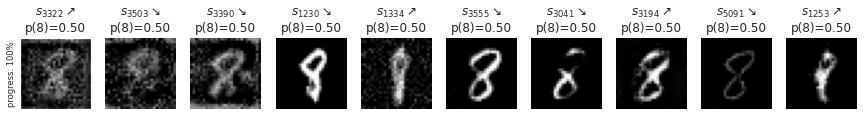}
\includegraphics[width=0.48\textwidth]{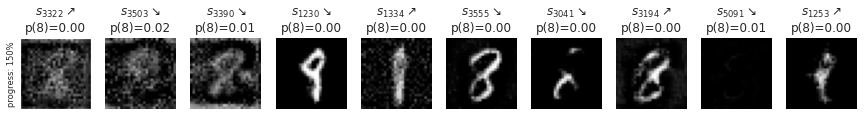}
\caption{Illustration of the degradation evolution starting from the same original image for the ten most influential dimensions. Each column represents one of the top ten influential dimensions; each line represents a different shift value (which also varies per dimension). More specifically, a shift reference value is defined for each dimension as the value that makes the classifier output equal to $0.50$ for the corresponding generated image, and each line represents a fraction of the dimension-specific shift reference value, written on the left as \emph{progress}. Above the images are displayed the StyleSpace dimension index, an arrow representing the direction to follow (augment or reduce the value), and the classifier output for the true class.}
\label{fig:manipulate}
\end{figure}

\begin{figure}[t]
    \centering
    \begin{subfigure}{0.48\textwidth}
    \includegraphics[width=\textwidth]{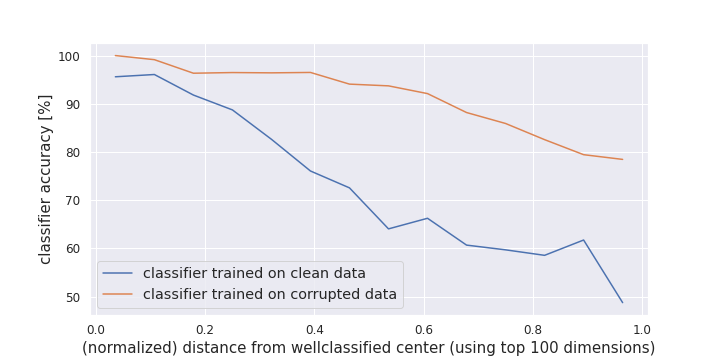}
    \caption{\label{fig:compare classifiers} Compare classifiers}
    \end{subfigure}
    \hfill
    \begin{subfigure}{0.48\textwidth}
    \centering
    \includegraphics[width=\textwidth]{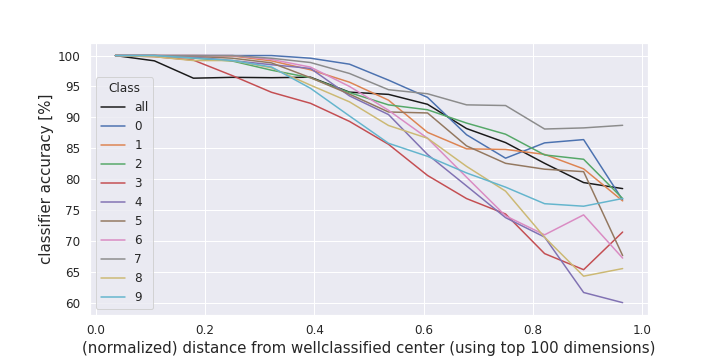}
    \caption{\label{fig:compare classes} Compare classes}
    \end{subfigure}
\caption{Classifier accuracy decreases when samples are far from the well-classified center in the latent space. \\
(\subref{fig:compare classifiers}) The classifier trained on clean data is less robust to corruptions: its accuracy decreases faster than the classifier trained on corrupted data. \\
(\subref{fig:compare classes}) (Using the classifier trained on corrupted data.) The robustness depends on the class. For instance, predicting correctly $3$, or $9$ is harder than $0$ or $7$. The lower start for the curve of \emph{all} classes can be explained by the fact that the well-classified center cannot work as well for all classes that for one class (we remind that the center and dimensions vary for each curve).
}
\label{fig:accuracy_vs_distance}
\end{figure}

\begin{figure}[t]
\centering
\includegraphics[width=0.48\textwidth]{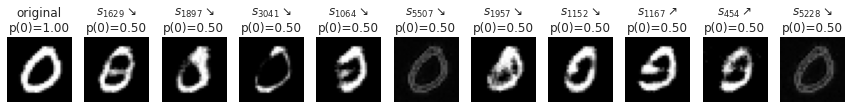}
\includegraphics[width=0.48\textwidth]{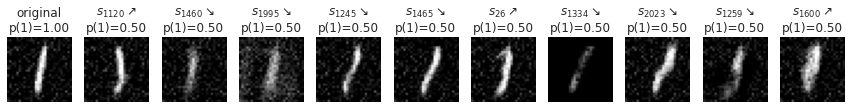}
\includegraphics[width=0.48\textwidth]{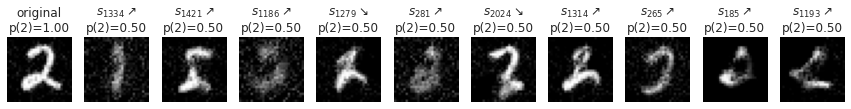}
\includegraphics[width=0.48\textwidth]{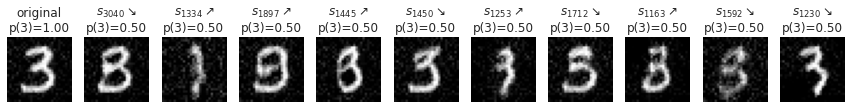}
\includegraphics[width=0.48\textwidth]{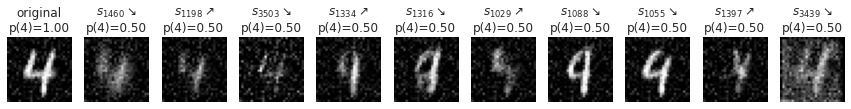}
\includegraphics[width=0.48\textwidth]{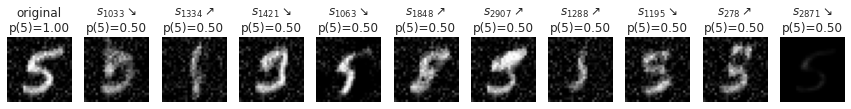}
\includegraphics[width=0.48\textwidth]{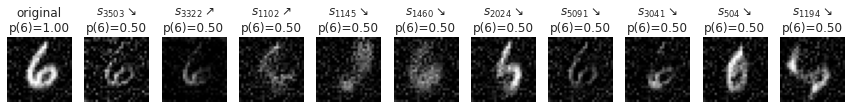}
\includegraphics[width=0.48\textwidth]{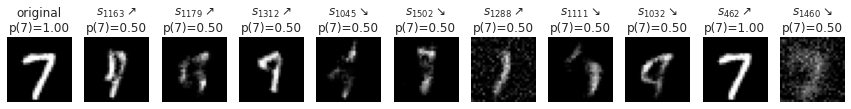}
\includegraphics[width=0.48\textwidth]{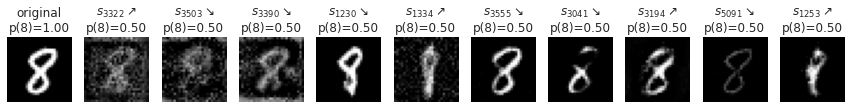}
\includegraphics[width=0.48\textwidth]{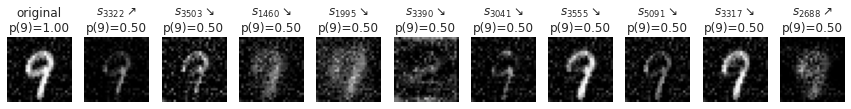}
\caption{Illustration of corner cases found for each class. By starting from the average image in StyleSpace for the class, we manipulate its latent code in one of the top ten influential dimensions until the classifier shifts its prediction (softmax probability for the given class falls to $0.50$). Above the images are displayed the StyleSpace dimension index, an arrow representing the direction to follow (augment or reduce the value), and the classifier output for the true class.}
\label{fig:corner_cases}
\end{figure}

\subsection{Implementation}

We used the MNIST dataset \cite{lecun2010mnist} to evaluate our approach. More precisely, we augmented the original data by introducing corruptions to simulate poor-quality data acquisition that may have an influence on class prediction. In particular, we chose Gaussian Noise and Gaussian Blur from \cite{imagenet-c} because they have a significant impact on classification accuracy for a classifier trained on clean data. Data are corrupted in the following way: the first half of the dataset remains clean, and the second half is first blurred (with a severity level randomly chosen between $1$, $2$ and $3$), and noise is added (with a severity level also randomly chosen between $1$, $2$ and $3$). It ensures that most of the samples remain visually recognizable. Random samples are shown in Figure~\ref{fig:samples_real_and_generated}.

The StyleGAN2 generative models contains three different latent spaces. It is generally admitted that the so-called StyleSpace $\mathcal{S}$ is a more disentangled representation space. We found that the well-classified and mis-classified samples are better separated in this space, even though the generation was not constrained in any way by the classifier. We can visualize this in Figure~\ref{fig:t-SNE} by using t-SNE projection \cite{t-SNE}.

After training on corrupted data, the classifier (a simple Convolutional Neural Network) reaches an accuracy of $97\%$ on the test data.
The metric used to quantify the performance of the generative model is the Fréchet Inception Distance (FID) \cite{FID}. The generative model trained on corrupted data reaches an FID of $1.63$ (computed by comparing $50k$ generated images, unfiltered and without using truncation, to the $60k$ images of the whole training dataset). This low value means a high generation quality. A few samples are shown in Figure~\ref{fig:samples_real_and_generated} demonstrating the capacity of the generative model to encode various levels and nature of corruption.

 \subsection{Influential dimensions}

We apply the method described in~\ref{sec:finding_dimensions} to rank dimensions in the learned StyleSpace. In order to verify that several dimensions have a bigger impact to performance than others, we computed the histograms of the two populations $S_{mis}$ and $S_{well}$ on each dimension. Figure~\ref{fig:histograms_s} depicts a selection of histograms. We  see that the values for the top dimensions follow different distributions for well-classified vs. mis-classified images, whereas random dimensions do not discriminate, meaning that the corresponding style attribute does not influence performance.

Figure~\ref{fig:manipulate} shows the impact of manipulating the latent codes by shifting values along the most influential dimensions. Each column represents one of the top ten influential dimension and each line represents a different shift value. We clearly observe various types of image corruption that can be interpreted a posteriori when increasing the shift value: the first three dimensions seem to introduce more  noise, dimensions 4, 5 and 10 deform the original shape, dimension 9 lowers the intensity, dimensions 6, 7 and 8 introduce partial occlusions. Using a generative model allows a large corruption vocabulary, and in particular allows shape deformation, a capacity that is not available in filter-based frameworks like Imagenet-C~\cite{imagenet-c}.

The last three lines of Figure~\ref{fig:manipulate} show images corresponding to a steep decrease of the classifier output score (from 1.0 to 0). This is where the class prediction shifts and where the generated image can be considered as a corner case (see section~\ref{sec:corner_cases}).

\subsection{Accuracy in the latent space}

As explained in Section~\ref{sec:accuracy}, we can look at the classifier accuracy evolution when fed with populations of various corruption levels sampled in the StyleSpace. Figure~\ref{fig:accuracy_vs_distance} shows this evolution on two different classifiers. The accuracy degradation is representative of the robustness to corruptions of a classifier: the classifier trained on clean data sees its accuracy decrease faster than the classifier trained on corrupted data. It also shows that it depends on the class: some classes are more difficult to predict than others. 

Using most or all dimensions of $\mathcal{S}$ to compute the distance makes the curve not monotonically decreasing. It is better to use fewer dimensions, e.g. $100$, as it makes the accuracy curve  decrease faster and monotonically. To make the curve clearer, we filtered out samples at too high distance values, where the generation quality decreases and the lower number of samples degrades the accuracy computation .

\subsection{Identification of corner cases}
\label{sec:corner_cases}

The identification of corner cases, as described in section \ref{sec:image_manipulation}, is illustrated for the $10$ classes in figure \ref{fig:corner_cases}. For each class, we illustrate the impact of the $10$ most influential dimensions separately, where the first column represents the average image of each class, and the ten following columns represent the result of corrupted image for one single dimension. Dimensions are selected among the most influential ones, by skipping those with no effect on the classification. It can be seen from the figure that visual results of corner cases are specific for each class. Then, if we compare the digits $4$ and $3$, we can see that corner cases of the digit $4$ are built by noise adding or structure deformation, whereas the construction of corner cases of the digit $3$ are characterized by class switching, into digits $8$, $1$ and $5$.

It is important to highlight that the obtained results from figure \ref{fig:corner_cases} are showing the impact of each single dimension by keeping all other dimensions in their optimal values (average image). The manipulation of a large set of dimensions could combine several types of degradation and allow the identification of new corner cases.


\section{Conclusion and perspectives}

The current work addresses a relationship between data quality and model performance by exploring the latent feature space of a generative model. Indeed, using our approach, we are able to identify the influential directions which deteriorate the classifier performances and discover corner cases in this space. The proposed approach is based on ranking the latent space dimensions using the classifier output gradient with respect to the StyleSpace input. 

Our results show the impact and the influence of each identified direction in terms of performance degradation on the classifier. These identified directions, separately or jointly, allow a visual account of the degradation which could help in the interpretability and explainability of deep learning classifiers.    


Despite the first promising conclusions of this work, our approach has been demonstrated only for generated and synthetic images. Its application to real data requires a capacity to encode -- or invert -- any data in the latent space \cite{xia2022gan}, to be able to apply the degradation encoded by the influential directions.

Another perspective, is to evaluate our approach on more complex data to identify other types of degradation attributes. Recent works on image manipulation show that visual attributes can be controlled for more complex images \cite{stylespace,StyleCLIP,GANSpace,EditGAN} and that generative models can be applied to larger datasets such as ImageNet \cite{sauer2022stylegan}. Those two advances indicate the possibility of scaling-up our approach.




\begin{acknowledgments}
    This work has been supported by the French government under the "Investissements d'avenir” program, as part of the SystemX Technological Research Institute. \\
    This work was granted access to the HPC/AI resources of IDRIS under the allocation 2022-AD011013372 made by GENCI.
\end{acknowledgments}

\bibliography{ijcai22,ijcai22_sh}


\end{document}